%% file: main_workshop.tex
\documentclass[conference]{IEEEtran}
\IEEEoverridecommandlockouts
\usepackage[pagebackref,breaklinks,colorlinks]{hyperref}
\usepackage{cite}
\usepackage{amsmath,amssymb,amsfonts}
\usepackage{algorithmic}
\usepackage{algorithm2e}
\usepackage{graphicx}
\usepackage{textcomp}
\usepackage{xcolor}
\usepackage{multirow}
\usepackage{makecell}
\usepackage{comment}
\usepackage{hhline}
\usepackage{tabularx}
\usepackage{url}
\def\BibTeX{{\rm B\kern-.05em{\sc i\kern-.025em b}\kern-.08em
    T\kern-.1667em\lower.7ex\hbox{E}\kern-.125emX}}

\usepackage{pifont}%
\newcommand{\xmark}{\ding{55}}%
\newcommand{\cmark}{\ding{51}}%

\begin{document}

\title{QClusformer: A Quantum Transformer-based Framework for Unsupervised Visual Clustering
}

\author{
Xuan-Bac Nguyen$^1$, Hoang-Quan Nguyen$^1$, Samuel Yen-Chi Chen$^2$\\
Samee U. Khan$^3$, Hugh Churchill$^4$, Khoa Luu$^1$\\
$^1$CVIU Lab, Dept. of EECS, University of Arkansas $\quad$ $^2$Wells Fargo $\quad$ \\
$^3$Mississippi State University $\quad$ $^4$ Dept. of Physics,  University of Arkansas\\
\small \texttt{\{xnguyen, hn016, hchurch, khoaluu\}@uark.edu, yen-chi.chen@wellsfargo.com, skhan@ece.msstate.edu} \\
\tt\small\url{https://uark-cviu.github.io/} 
}

\maketitle

\begin{abstract}
Unsupervised vision clustering, a cornerstone in computer vision, has been studied for decades, yielding significant outcomes across numerous vision tasks. However, these algorithms involve substantial computational demands when confronted with vast amounts of unlabeled data. Conversely, quantum computing holds promise in expediting unsupervised algorithms when handling large-scale databases. In this study, we introduce QClusformer, a pioneering Transformer-based framework leveraging quantum machines to tackle unsupervised vision clustering challenges. Specifically, we design the Transformer architecture, including the self-attention module and transformer blocks, from a quantum perspective to enable execution on quantum hardware.
In addition, we present QClusformer, a variant based on the Transformer architecture, tailored for unsupervised vision clustering tasks. By integrating these elements into an end-to-end framework, QClusformer consistently outperforms previous methods running on classical computers. Empirical evaluations across diverse benchmarks, including MS-Celeb-1M and DeepFashion, underscore the superior performance of QClusformer compared to state-of-the-art methods.
\end{abstract}


\section{Introduction}

Quantum computing derived from quantum mechanics can exponentially accelerate the solution of specific problems compared to classical computing, thanks to the unique properties of superposition and entanglement \cite{preskill2012Quantum,preskill2018Quantum,boixo2018characterizing}.
As the number of available quantum bits, i.e., qubits, is increasing in the noisy intermediate-scale quantum (NISQ) era, quantum computing can achieve potential advantages.
Quantum Machine Learning (QML) is one of the most popular applications in quantum computing because it requires computing power and is robust to noise. In recent years, many studies have been presented to develop QML frameworks that are equivalent to classical ones, such as quantum k-nearest neighbor \cite{basheer2020Quantum}, quantum support vector machines \cite{rebentrost2014Quantum}, quantum clustering \cite{horn2001method,horn2001algorithm,nguyen2023Quantum}, and quantum neural networks (QNNs) \cite{ezhov2000Quantum,zhou2023Quantum,gupta2020Quantum, Dendukuri2019b, Dendukuri2019a}.
Because of its quantum properties, QML has the potential computational and storage efficiency compared to classical machine learning \cite{biamonte2017Quantum,du2020expressive}.
It leads to the ability to solve machine learning problems with large-scale datasets using quantum computing.

Unsupervised clustering is a fundamental paradigm where algorithms autonomously uncover patterns and structures within data without explicit guidance from labeled examples. Unlike supervised learning, which relies on labeled data for training, unsupervised clustering operates on unlabeled data, making it exceptionally versatile for tasks where labeled data is scarce or expensive. This autonomous learning capability renders unsupervised clustering indispensable across various domains, including but not limited to natural language processing, computer vision, etc. In the typical approach, since the classical clustering methods, i.e., K-means \cite{lloyd1982least,sculley2010web}, and DBSCAN \cite{ester1996density}, iteratively assign the samples into clusters based on the distance metrics, they are computationally expensive to apply in large-scale datasets. This limitation poses a significant challenge for clustering research. However, the superposition and entanglement features of quantum computing enable quantum computers to store and process extensive amounts of data. These capabilities light the way for exploring machine learning algorithms, i.e., unsupervised clustering on quantum platforms, opening up exciting avenues for research.

\begin{figure*}[t]
    \centering
    \includegraphics[width=0.9\linewidth]{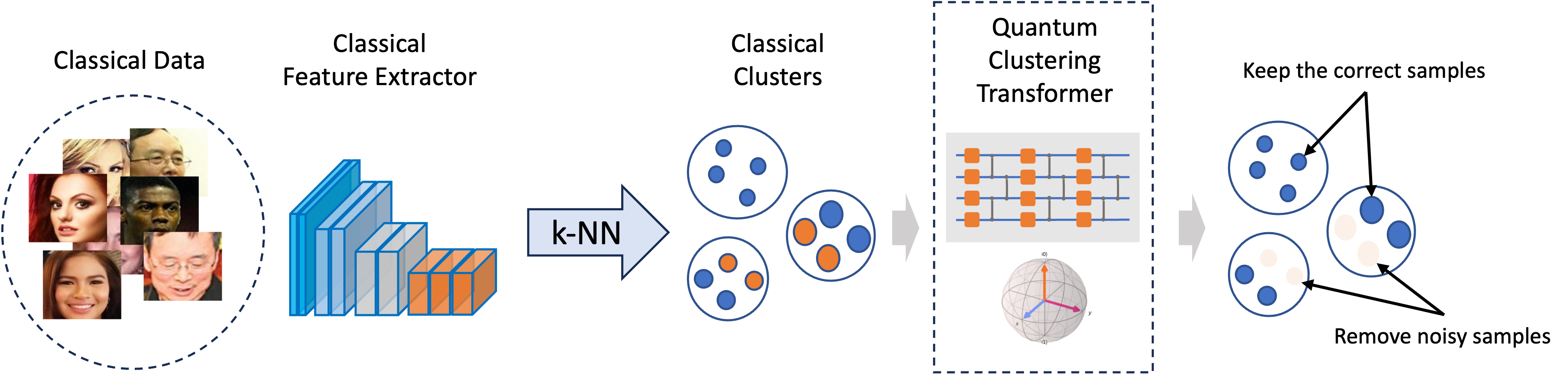}
    \caption{
    \textbf{An overview of the Quantum Transformer for Clustering framework.} 
    Given classical data, i.e., images, the samples are extracted into feature vectors via a classical deep learning model.
    Then, a k-nearest neighbor algorithm is applied to cluster the samples.
    To automatically select the correct samples in each cluster, we propose a novel Quantum Clustering Transformer (QClusformer) justifying the correlation between feature vectors.
    }
    \label{fig:overview}
\vspace{-4mm}
\end{figure*}

Inspired by the unique properties of quantum machines, this paper presents an innovative quantum clustering framework utilizing Transformer architecture to autonomously cluster samples in an unsupervised fashion. The approach demonstrates resilience in handling noisy and challenging samples, owing to its efficient self-attention mechanism. In addition, leveraging the benefits of quantum computing, the method offers ease of optimization. The framework of the Quantum Transformer for Clustering is illustrated in Fig. \ref{fig:overview}.

The contributions of this work are threefold. Initially, we propose a Transformer-based method, named QClusformer, for visual clustering within the realm of quantum computing. Subsequently, we introduce a self-attention module and Transformer layers employing parameterized quantum circuits. Lastly, the empirical experimental results on various benchmark datasets demonstrate the novelties and efficiency of our proposed approach compared to the corresponding classical machines.

\section{Background}
\label{sec:background}

\subsection{Parameterized Quantum Circuit}

The parameterized quantum circuit (PQC) \cite{benedetti2019parameterized} is a special kind of quantum circuit with parameters that can be optimized or learned iteratively.
The PQC comprises three parts: data encoding, parameterized layer, and quantum measurements.

Given a classical data $\mathbf{z} \in \mathbb{R}^{D}$ where $D$ is the data dimension, the data encoding circuit $U(\mathbf{z})$ is used to transform $\mathbf{z}$ into a quantum state $|\psi\rangle$.
The quantum state $|\psi\rangle$ is transformed via the parameterized layer $V(\theta)$ to a new state $|\psi\rangle$.
The parameterized layer is a sequence of quantum circuit operators with learnable parameters denoted as:
$V(\theta) = V_L(\theta_L) V_{L-1}(\theta_{L-1}) \dots V_{1}(\theta_{1})$,
where $L$ is the number of operators.
The learnable parameters can be updated via gradient-based \cite{mitarai2018Quantum}, or gradient-free \cite{chen2022variational} algorithms.
The quantum measurements $H$ are used to retrieve the values of the quantum state for further processing.
Overall, the PQC can be formulated as:
\begin{equation}
    \langle H \rangle = \langle 0 | U^\dagger(\mathbf{z}) V^\dagger(\theta) H V(\theta) U(\mathbf{z}) | 0 \rangle
\end{equation}
where $H$ is a predefined observable.

PQC uses a hybrid quantum-classical procedure to optimize the trainable parameters iteratively.
All learning methods take the training data as input and evaluate the model performance by comparing the predicted and ground-truth labels. 
Based on this evaluation, the methods update the model parameters for the next iteration and repeat the process until the model converges and achieves the desired performance. 

\begin{figure}[t]
    \centering
    \includegraphics[width=0.95\linewidth]{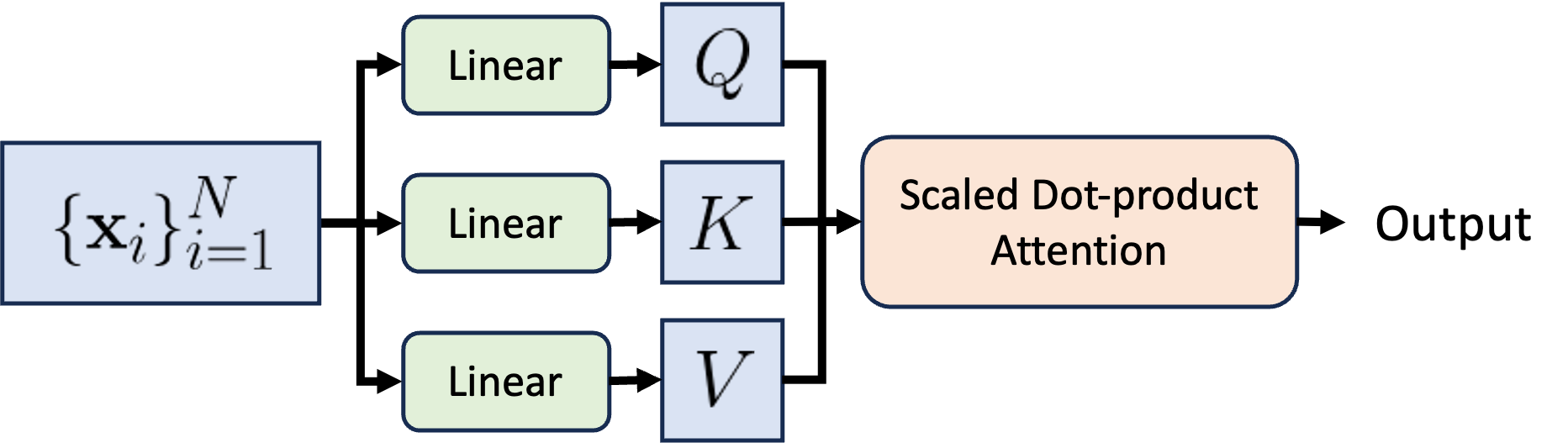}
    \caption{A framework of the self-attention module on classical data.}
    \label{fig:classical_self_attention}
\vspace{-4mm}
\end{figure}

\subsection{Visual Clustering}

Given a dataset $\mathcal{D} = \{\mathbf{x}_1, \mathbf{x}_2, \dots \mathbf{x}_n\}$ where $\mathbf{x}_i$ represents a data point in $d-$dimensional space, the objective of unsupervised clustering is to partition the dataset into $k$ clusters such that satisfies following conditions. 
Firstly, each data point belongs to exactly one cluster. 
Secondly, data points within the same cluster are more similar than those in different clusters. 
Depending on the predefined problem, the number of clusters $k$ can be determined automatically or specified beforehand. 
The goal is to find an optimal partitioning of the dataset that maximizes the intra-cluster similarity and minimizes the inter-cluster similarity. 
It is typically achieved by defining a suitable objective function or similarity measure and optimizing it to obtain cluster assignments. 
Formally, let $C = \{C_1, C_2, \dots ,C_k\}$ represent the set of $k$ clusters, where $C_i$ denotes the $i^{th}$ cluster, and $\mathbf{c}_i$ represents for the centroid (or center) of the $C_i$. 
The task is to find an optimal partitioning that minimizes the following objective function, quantifying the dissimilarity between data points and their respective cluster centroids: 
$J = \sum_{i=1}^{k}\sum_{\mathbf{x}\in C_i} dist(\mathbf{x}, \mathbf{c}_i)$, 
where $dist(\mathbf{x}, \mathbf{c}_i)$ denotes the distance between sample $\mathbf{x}$ and centroid $\mathbf{c}_i$ of the cluster $C_i$. Common distance measures used in unsupervised clustering include Euclidean distance, Manhattan distance, cosine similarity, etc. The optimization process typically involves iterative procedures such as K-means, hierarchical clustering, density-based clustering, or spectral clustering to find the optimal partitioning that minimizes the objective function.

\begin{figure*}[ht]
    \centering
    \includegraphics[width=0.8\linewidth]{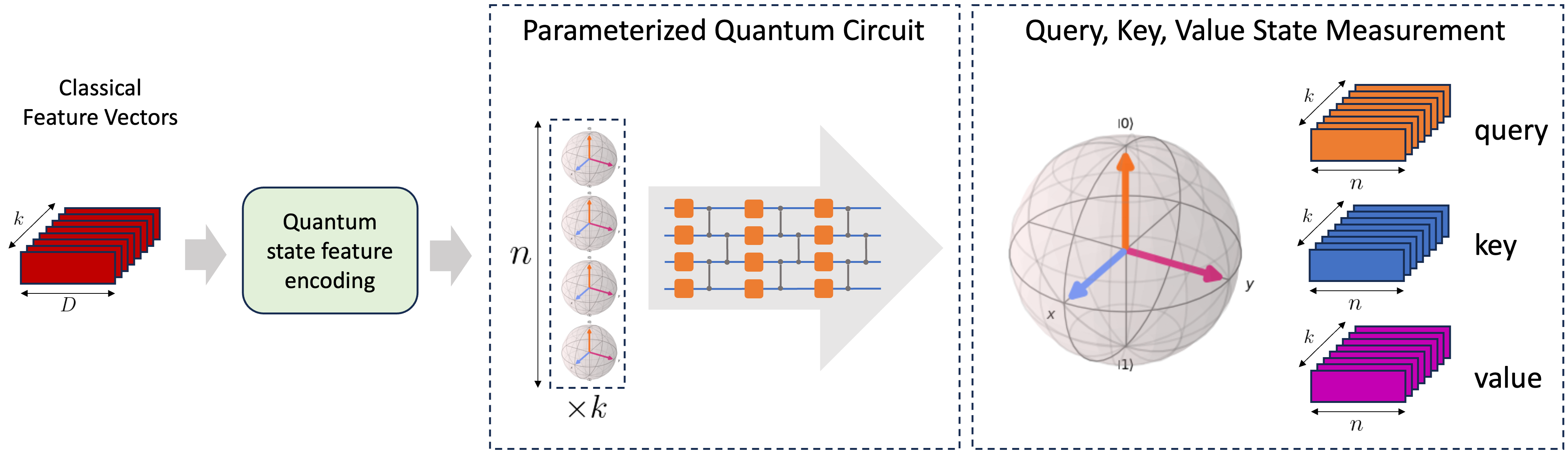}
    \caption{
    \textbf{The Quantum Self-attention Module.}
    Given $k$ encoded classical feature vectors sized $D$ of a cluster, we encode the feature vectors into $k$ quantum states.
    Each quantum state uses $n = \lceil\log_2(D)\rceil$ qubits to contain the information of the classical feature vector.
    After being transformed via a Parameterized Quantum Circuit, each quantum state is measured to obtain the query, key, and value for self-attention. 
    }
    \label{fig:QuantumTransformer}
\vspace{-4mm}
\end{figure*}

\subsection{Transformer}
\label{sec:background_transformer}

The Transformer \cite{vaswani2017attention} is a sequence-to-sequence model consisting of an encoder and a decoder. 
Each encoder and decoder is a stack of $T$ identical blocks.
Each block is composed of a multi-head attention module and a feed-forward network.
With the multi-head attention module, the Transformer can focus on different positions of an input sequence, leading to a robust deep network that outperforms prior methods.

In each self-attention layer, the input data 
$\{\mathbf{x}_i \in \mathbb{R}^d\}_{i=1}^{N}$ are linearly mapped via three weighted matrices, i.e., key $W_k \in \mathbb{R}^{d \times d}$, query $W_q \in \mathbb{R}^{d \times d}$, and value $W_v \in \mathbb{R}^{d \times d}$, to three parts $\mathbf{k}_i = W_k \mathbf{x}_i$, $\mathbf{q}_i = W_q \mathbf{x}_i$, and $\mathbf{v}_i = W_v \mathbf{x}_i$, respectively.
The query and key parts are then applied to the inner product, and the output is computed as:
$\mathbf{y}_i = \sum_{j=1}^{N} a_{i,j} \mathbf{v}_j$, 
where
$a_{i,j} = \frac{\exp(\mathbf{q}_i^\top \mathbf{k}_j)}{\sum_{l=1}^{N} \exp(\mathbf{q}_i^\top \mathbf{k}_l)}$
is the self-attention coefficient. 
Fig. \ref{fig:classical_self_attention} illustrates the framework of the self-attention module on classical data.

\section{Our Proposed Approach}
\label{sec:approach}

\subsection{Motivations}

Assume that we have employed a clustering algorithm, i.e., K-Nearest Neighbors (KNN), to form initial clusters $C_i$. Given a cluster $C_i$ of $k$ samples with a center $\mathbf{c}_i$. This cluster may include erroneous samples due to various factors. Firstly, setting a fixed number of neighbors, denoted as $k$, can result in clusters with fewer than $k$ samples, potentially containing noisy or inaccurate data points. Secondly, the lack of robustness in the feature extractor, such as a deep neural network, may cause the representations of distinct samples to be similar, resulting in samples from different subjects being differently grouped into the same cluster. Lastly, the unlabeled data may introduce challenging samples that closely resemble each other, causing misassignments in cluster allocation when employing K-NN.

Numerous methodologies have tackled this problem on classical computers, employing a variety of techniques such as graph-based approaches \cite{wang2019linkage, yang2020learning, yang2019learning, shen2021structure, shen2023clip, shin2023local}, and transformer-based methods \cite{nguyen2021clusformer}. Although transformer architectures have shown remarkable success in diverse computer vision tasks \cite{li2022blip, yu2022coca, zhai2023sigmoid, luo2023lexlip, wang2023equivariant, nguyen2023micron, nguyen2023insect, nguyen2020self, nguyen2019audio,nguyen2019sketch,nguyen2022multi,nguyen2022two,nguyen2023algonauts,nguyen2023brainformer,nguyen2023fairness, nguyen2024diffusion, nguyen2024quantum}, their potential within quantum computing remains promising.
Hence, leveraging insights from classical Transformers, we expand its application to quantum machines by conceptualizing the most important element of the Transformer, self-attention, through a quantum perspective. With this concept, we construct successive Transformer blocks comprising self-attention modules and Parameterized Quantum Circuits (PQC) serving as linear layers to build a quantum transformer architecture. 
We propose a Transformer-based clustering framework for quantum machines to recognize the noisy or hard samples inside the cluster $C_i$.

\subsection{Quantum State Feature Encoding}

To operate on a classical dataset for a quantum circuit, it is important to define an encoding method to transform the classical feature vector into a quantum state. As the number of qubits is limited, we need to encode the classical values with a small number of qubits as much as possible.
Given $n$ qubits, a general quantum state can be represented as:
\begin{equation}
    \left| \psi \right> = \sum_{(q_1, \dots, q_n) \in \{0, 1\}^n}
    c_{q_1, \dots, q_n} | q_1 \rangle \otimes \dots \otimes | q_n \rangle
\end{equation}
where $c_{q_1, \dots, q_n} \in \mathbb{C}$ is the amplitude of a quantum state.

In this work, we use the amplitude encoding method to encode the classical vector into a quantum state.
Given a feature vector $\mathbf{s} \in \mathbb{R}^{D}$ where $D \geq 2$, the quantum state can be represented as: 
$|\psi\rangle = \sum_{i=1}^{D} s_i | i - 1 \rangle$.
Indeed, to encode a $D$-dimension feature vector, the minimum number of required qubits is $n = \lceil \log_2(D) \rceil$.
The encoded quantum state can then be transformed by quantum gates and then measured to a classical $n$-dimension vector for the quantum self-attention in a lower space dimension. 

\subsection{Quantum Transformer}

In contrast to classical transformers, which employ linear transformations as detailed in Section \ref{sec:background_transformer}, PQCs transform classical features for self-attention. Given a sequential input $S = {\mathbf{s}^{(h)}}_{h=1}^k \in \mathbb{R}^{k \times D}$, quantum self-attention is introduced to compute the relationship between feature vectors in $S$ as follows:
\begin{equation}
\begin{split}
    Q &= \text{PQC}_{Q}(S) \in \mathbb{R}^{k \times n}, \quad
    K = \text{PQC}_{K}(S) \in \mathbb{R}^{k \times n}, \\
    V &= \text{PQC}_{V}(S) \in \mathbb{R}^{k \times n}, \quad
    Z^\prime = \text{softmax}\left(\frac{Q K^\top}{\sqrt{n}}\right) V
\end{split}
\end{equation}

In general, when provided with a $D$-dimensional vector $\mathbf{s}$, our approach involves encoding the values into a quantum state $|\psi\rangle$ through an $n$-qubit circuit. Following this, the quantum state transforms a parameterized layer. Subsequently, the final state is measured for subsequent processing, typically utilizing the Pauli $Z$ matrix as an observable in most applications. However, this design utilizes three parameterized quantum circuits to compute the key, query, and value of the feature vectors, potentially not fully leveraging all the information encoded in the quantum state.

Moreover, we found that given an arbitrary 1-qubit quantum state $|\psi\rangle$, the density matrix $\rho = |\psi\rangle\langle\psi|$ can be expressed as a linear combination of Pauli matrices:
\begin{equation}
\begin{split}
    \rho &= \frac{1}{2}\left( I + r_x \sigma_x + r_y \sigma_z + r_z \sigma_z \right)
\end{split}
\end{equation}
where $r_i \in \mathbb{R}$ and $r_x^2 + r_y^2 + r_z^2 = 1$. 
Then, for each observable $\sigma_i \in \{ \sigma_x, \sigma_y, \sigma_z \}$, the expectation of the measurement is:
\begin{equation}
\begin{split}
    \langle \sigma_i \rangle &= \text{Tr}\left( \frac{1}{2}\left( I + r_x \sigma_x + r_y \sigma_z + r_z \sigma_z \right) \sigma_i \right) = r_i
\end{split}
\label{eq:measurement_xyz}
\end{equation}
Eqn. \ref{eq:measurement_xyz} shows that the three Pauli matrices as observables measure three separated values of the quantum state $\rho$.
Inspired by this property, we design a single parameterized quantum circuit for the self-attention module as shown in Fig. \ref{fig:QuantumTransformer}.
Each observable computes the key, query, or value of the feature vector.
Given an $n$-qubit parameterized quantum circuit, the quantum self-attention computes $n$-dimension key, query, and value.

Let a QClusformer encoder $E(S)$ be a stack of $T$ Quantum Transformer blocks where each block contains a quantum self-attention (QSA) and a parameterized quantum layer (PQL) 
:
\begin{equation}
\begin{split}
    Z^{\prime(t)} &= Z^{(t-1)} + \text{QSA}(\text{LN}(Z^{(t-1)})) \\
    Z^{(t)} &= Z^{\prime(t)} + \text{PQL}(\text{LN}(Z^{\prime(t)})) \\
    Z^{(0)} &= S, \quad 1 \leq t \leq T
\end{split}
\end{equation}
where $\text{LN}$ is the layer normalization.
The output of the QClusformer encoder $E$ is used for the clustering task.

\section{Implementation}
\label{sec:implementation}

\subsection{Visual Cluster Dataset}

Given a set of features $\mathcal{F} = \{\mathbf{f}_i\}_{i=1}^{N}$ extracted from dataset $\mathcal{D}$, the k-nearest neighbors $\mathcal{K}$ is applied to cluster the samples based on the cosine similarity score.
They will be formed as a cluster $C_i$ having $\mathbf{c}_i$ as the center:
\begin{equation}
    C_i = \mathcal{K}(\mathbf{c}_i, \mathcal{F}, k) \in \mathbb{R}^{k \times D}
\end{equation}
where $k$ is the number of nearest neighbors.
Then, we construct a clustered dataset denoted as $C = \{C_i\}_{i=1}^{N}$.
This dataset is used in the QClusformer, which will be described in the following sections.

\subsection{Cosine Similarity Encoding}

As the Transformer processes the input represented as a sequence, the cluster $\mathbf{C}_i$ has to be represented as a sequence $S_i = \mathcal{G}(C_i)$ for QClusformer.
Generally, a Transformer encoder expects an input of a sequence similar to Recurrent Neural Networks or Long-Short Term Memory that is widely used in Natural Language Processing. 
As the Transformer processes words in parallel, a Positional Encoding is presented to preserve the order of the sequence. 
Unlike the sequence inputs, the cluster dataset follows the similarities between samples and the cluster centers.
Thus, a Cosine Similarity Encoding is introduced to describe the structure of the cluster dataset.
Let $h$ be the position of an element in the sequence input, $\mathbf{e}_i^{(h)} \in \mathbb{R}^{k}$ be the Cosine Similarity Encoding as follows,
\begin{equation}
    \mathbf{e}_i^{(h)} = \{ \text{similarity}(\mathbf{c}_i, \mathbf{f}_i^{(j)}) \}_{j=1}^{k}
\end{equation}
The feature of the $h$-th element in the sequence turns to
\begin{equation}
    \mathbf{s}_i^{(h)} = \mathbf{f}_i^{(h)} + \mathbf{e}_i^{(h)} W_e
\end{equation}
where $\mathbf{W}_e \in \mathbb{R}^{k \times D}$ is a projected weight.
The feature vector $\mathbf{s}_i^{(h)}$ is then encoded into a quantum state for the quantum self-attention.

\subsection{Objective and Loss Functions}

Although the neighbors of center $\mathbf{c}_i$ are expected to have the same label as the center, there are hard samples from various clusters in real-world conditions.
Thus, $C_i$ cannot contain all visual samples in one label.
The QClusformer is introduced to detect these hard samples.
The output sequence is a binary sequence $y_i$, where $y_i^{(h)} = 1$ if the $h$-th sample in the sequence has the same label as the center $\mathbf{c}_i$ and vice versa.
Let $\hat{y}_i^{(h)}$ be the ground truth of the output, a Binary Cross Entropy loss is used to train the QClusformer: 
\begin{equation}
\small
    \mathcal{L}_i(y_i, \hat{y}_i) = - \sum_{h=1}^{k}[
    \hat{y}_i^{(h)} \log(y_i^{(h)}) + (1 - \hat{y}_i^{(h)}) \log(1 - y_i^{(h)})
    ]
\end{equation}

\section{Experimental Results}
\label{sec:experiment}

\subsection{Evaluation Metrics}

To evaluate the performance of the methods, we use Fowlkes Mallows Score to measure the similarity between two clusters with a set of points.
This score is computed by taking the geometry mean of precision and recall of the point pairs.
Thus, Fowlkes Mallows Score is also called Pairwise F-score ($F_P$), defined as: $F_P = \frac{TP}{\sqrt{(TP + FP) \times (TP + FN)}}$
where $TP$ is the number of point pairs in the same cluster in both ground truth and prediction,
$FP$ is the number of point pairs in the same cluster in prediction but not in ground truth,
and $FN$ is the number of point pairs in the same cluster in ground truth but not prediction.
BCubed F-score is another popular metric for clustering evaluation focusing on each data point.

\input{tables/ms1m}

\subsection{Datasets}

\begin{figure}[t]
    \centering
    \includegraphics[width=0.8\linewidth]{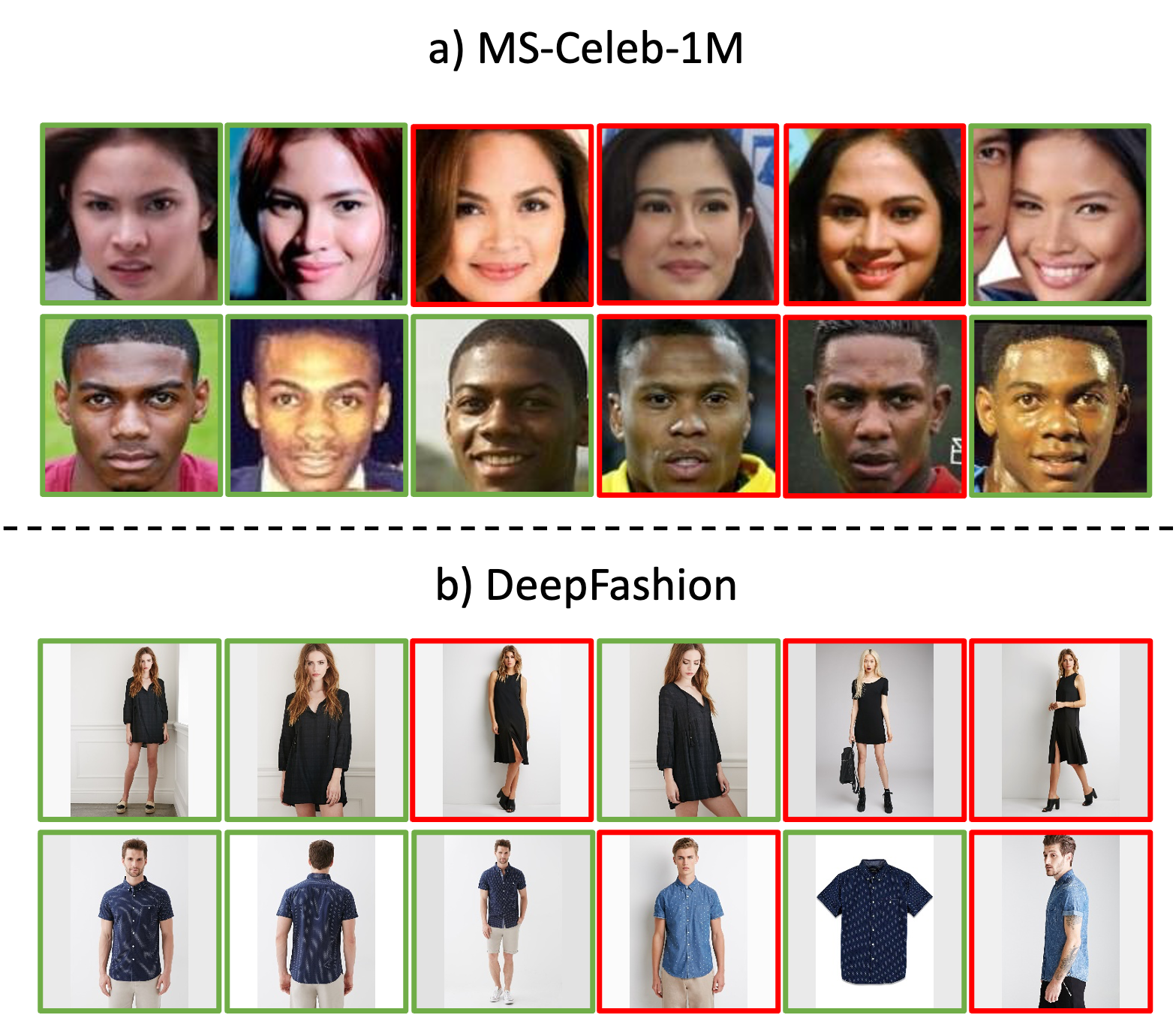}
    \caption{
    The MS-Celeb-1M and DeepFashion datasets are illustrated through samples. Each row represents either a subject. The first image in each row denotes the center of a cluster, while the subsequent images are the nearest neighbors of the first one, identified through the k-NN algorithm utilizing quantum features. Images bordered in red signify that they belong to a different class than the first image in the row, whereas those bordered in green share the same class as the first image.
    \textbf{Best view in color.}
    }
    \label{fig:dataset_examples}
\vspace{-4mm}
\end{figure}

We follow \cite{yang2020learning} to use MS-Celeb-1M \cite{guo2016ms} and DeepFashion \cite{liu2016deepfashion} datasets for experiments.
The Fig. \ref{fig:dataset_examples} illustrates samples of the datasets.

\noindent
\textbf{Face Clustering and Recognition.}
MS-Celeb-1M \cite{guo2016ms} is a large-scale face recognition dataset crawled from the Internet.
The cleaned version consists of 85K identities with 5.8M images. The images are pre-processed by aligning and cropping to the size of $112 \times 112$. The MS-Celeb-1M dataset is randomly split into ten parts. Each part contains approximately 584K images of 8,500 identities. 
There is no identity overlapped among them.

\noindent
\textbf{Clothes Clustering.}
DeepFashion \cite{liu2016deepfashion} is a large-scale clothes recognition dataset.
Inspired by Yang et al. \cite{yang2020learning} for clustering, the DeepFashion dataset is split into 25,752 images of 3,997 categories for training and 26,960 images of 3,984 categories for testing. Similar to the face clustering setting, no overlapped category exists between the training and testing sets.

\subsection{Performance on MS-Celeb-1M Clustering}

Table \ref{tab:experiment_ms1m} illustrates the performance of our proposed method on the MS-Celeb-1M Clustering benchmark.
Following the benchmarks from the previous studies \cite{nguyen2021clusformer,wang2022ada,wang2019linkage,yang2019learning,yang2020learning}, we first train the classical feature extractor on the first part of the MS-Celeb-1M dataset in a supervised manner.
Then, we construct the cluster dataset using $k$-nearest neighbor and train the QClusformer model.
The trained QClusformer model is evaluated on five accumulated parts as shown in Table \ref{tab:experiment_ms1m}.
Due to hardware constraints, we can only simulate the QClusformer with the number of Transformer blocks $T=1$.
For a fair comparison, we train the classical Clustering Transformer with the same number of blocks.
Compared to the classical Clustering Transformer, the proposed QClusformer achieves higher performance with a Pairwise F1-score from $63.31\%$ to $74.50\%$ and BCubed F1-score from $79.74\%$ to $82.09\%$ on the 584K testing part.
Similar performance trends are shown across all testing parts of MS-Celeb-1M. %

\subsection{Performance on DeepFashion Clustering}

\input{tables/deepfashion}

We compare the performance of our proposed method on the DeepFashion dataset for the clustering task as shown in Table \ref{tab:experiment_deepfashion}.
The evaluation protocols on the DeepFashion clustering task are the same as the MS-Celeb-1M clustering benchmark.
While achieving higher performance than the previous classical clustering methods with a Pairwise F1-score of $35.71\%$ and BCubed F1-score of $60.00\%$, the proposed QClusformer method maintains competitive results compared to the classical Transformer setting.

\section{Conclusion}
\label{sec:conlusion}

In this paper, we have proposed QClusformer, a quantum approach, for Transformer-based vision clustering problems.
The QClusformer method utilizes Parameterized Quantum Circuits to leverage the quantum information in self-attention computing.
By separating the values of a quantum state via Pauli matrices for measurement, we minimize the usage of quantum computing resources for the self-attention module.
The competitive evaluation results on multiple large-scale vision clustering benchmarks have demonstrated the potential of the proposed Transformer-base clustering framework on quantum computing in various applications.

{
    \small
    \bibliographystyle{IEEEtran}
    \bibliography{main,bib/qml_example}
}

\end{document}

%% file: tables/ms1m.tex
\begin{table*}[t]
\centering
\caption{Experimental results of face clustering w.r.t the different number of unlabelled test
sets}
\begin{tabular}{l|c|cc|cc|cc|cc|cc}
\Xhline{2\arrayrulewidth}
\multirow{2}{*}{Method} & Num. unlabeled & \multicolumn{2}{c}{584K} & \multicolumn{2}{c}{1.74M} & \multicolumn{2}{c}{2.89M} & \multicolumn{2}{c}{4.05M} & \multicolumn{2}{c}{5.21M} \\
& QML & $F_P$   & $F_B$   & $F_P$   & $F_B$   & $F_P$   & $F_B$   & $F_P$   & $F_B$   & $F_P$   & $F_B$   \\
\hline
K-means \cite{lloyd1982least,sculley2010web} & \xmark & 79.21 & 81.23 & 73.04 & 75.20 & 69.83 & 72.34 & 67.90 & 70.57 & 66.47 & 69.42 \\
HAC \cite{sibson1973slink}    & \xmark & 70.63 & 70.46 & 54.40 & 69.53 & 11.08 & 68.62 & 1.40  & 67.69 & 0.37  & 66.96 \\
DBSCAN \cite{ester1996density} & \xmark & 67.93 & 67.17 & 63.41 & 66.53 & 52.50 & 66.26 & 45.24 & 44.87 & 44.94 & 44.74 \\
ARO \cite{otto2017clustering}    & \xmark & 13.60 & 17.00 & 8.78  & 12.42 & 7.30  & 10.96 & 6.86  & 10.50 & 6.35  & 10.01 \\
CDP \cite{zhan2018consensus}    & \xmark & 75.02 & 78.70 & 70.75 & 75.82 & 69.51 & 74.58 & 68.62 & 73.62 & 68.06 & 72.92 \\
\hline
Classical Clusformer & \xmark & 63.31 & 79.74 & 60.23 & 78.14 & 58.53 & 76.84 & 56.79 & 75.92 & 55.13 & 75.01 \\
\textbf{QClusformer}          & \cmark & \textbf{74.50} & \textbf{82.09} & \textbf{73.12} & \textbf{80.92} & \textbf{71.25} & \textbf{78.67} & \textbf{69.46} & \textbf{77.24} & \textbf{68.38} & \textbf{75.86}\\
\Xhline{2\arrayrulewidth}
\end{tabular}
\label{tab:experiment_ms1m}
\vspace{-4mm}
\end{table*}

%% file: tables/deepfashion.tex
\begin{table}[t]
\centering
\caption{Experimental Results of DeepFashion Clustering.}
\begin{tabular}{l|c|cc}
\Xhline{2\arrayrulewidth}
Method               & QML & $F_P$ & $F_B$ \\
\hline
K-means \cite{lloyd1982least,sculley2010web} & \xmark             & 32.86                   & 53.77                   \\
HAC \cite{sibson1973slink} & \xmark                & 22.54                   & 48.77                   \\
DBSCAN \cite{ester1996density} & \xmark             & 25.07                   & 53.23                   \\
Mean shift \cite{comaniciu2002mean} & \xmark           & 31.61                   & 56.73                   \\
Spectral \cite{ng2001spectral} & \xmark           & 29.02                   & 46.40                   \\
\hline
Classical Clusformer & \xmark & 35.62                   & 60.61                   \\
\textbf{QClusformer}          & \cmark & \textbf{35.71}                   & \textbf{60.00}                   \\
\Xhline{2\arrayrulewidth}
\end{tabular}
\label{tab:experiment_deepfashion}
\vspace{-4mm}
\end{table}